\documentclass[11pt]{article}

\usepackage[final]{acl}

 \usepackage{microtype}
\usepackage{graphicx}
\usepackage{booktabs}
\usepackage{amsmath}
\usepackage{amssymb}
\usepackage{multirow}
 \usepackage{times}
\usepackage{latexsym}
\usepackage[T1]{fontenc}
\usepackage[utf8]{inputenc}
\usepackage{makecell}

\usepackage[dvipsnames]{xcolor}
\everydisplay{\footnotesize}

\usepackage{tikz}
\usetikzlibrary{positioning,calc,fit,arrows.meta,backgrounds}

\usepackage[english,bidi=default]{babel} 
\babelfont{rm}{TeXGyreTermesX} 
\babelprovide[import]{hindi}
\babelfont[*devanagari]{rm}{Lohit Devanagari}
\babelprovide[import]{arabic}

\definecolor{maskgrey}{RGB}{226,226,230}
\definecolor{maskline}{RGB}{138,138,146}
\definecolor{masktext}{RGB}{92,92,100}
\definecolor{keepblue}{RGB}{204,223,244}
\definecolor{keepline}{RGB}{45,105,175}
\definecolor{keeptext}{RGB}{22,64,120}
\definecolor{tmplbg}{RGB}{253,235,211}
\definecolor{tmplline}{RGB}{197,133,32}
\definecolor{tem}{RGB}{203,238,255}
\definecolor{mypink}{RGB}{255, 222, 230}

\definecolor{cLine}{HTML}{3B4252}
\definecolor{cIO}{HTML}{ECEFF2}   \definecolor{cIOd}{HTML}{6B7280}
\definecolor{cPW}{HTML}{DDE8F4}   \definecolor{cPWd}{HTML}{31659B}
\definecolor{cMC}{HTML}{FBE7D6}   \definecolor{cMCd}{HTML}{B2701F}
\definecolor{cOP}{HTML}{E4EEE2}   \definecolor{cOPd}{HTML}{4E7A45}
\definecolor{cBG}{HTML}{F7F8FA}

\title{alkhder at ImageEval 2026 Shared Tasks: Constrained Selection and Rotation-Averaged Logit Scoring  for Multimodal Hallucination Detection}

\author{
  Maher Pasa \\
  Altınbaş University, T\"urkiye \\
  \small\texttt{engmaherpasa@gmail.com}
  \And
  Maria Hanini \\
  Imperial College London \\
  \small\texttt{m.hanini@imperial.ac.uk}
  \AND
  Amro Najjar \\
  Luxembourg Institute of Science and Technology \\
  \small\texttt{amro.najjar@list.lu}
  \And
  Hasan Alkhder \\
  Sakarya University, T\"urkiye \\
  \small\texttt{hasan.alkhder2@ogr.sakarya.edu.tr}
}

\begin{document}

\maketitle
\begin{abstract}

We describe our submission to Subtask~1b of ImageEval 2026. The task consists of selecting the single visually valid statement from three candidates paired with an image, and is in both English and Modern Standard Arabic (MSA). Our system uses a pre-trained 7B vision--language model in a zero-shot setting, without task-specific training or fine-tuning. We formulate the task as a constrained selection problem and choose the answer using next-token log-probabilities. To reduce answer-position bias, we apply a Latin-square rotation to the candidate statements. Our system achieves CI scores of 0.0510 on the English test set and 0.0960 on the MSA test set, reducing CI relative to the official 3B baselines reported on devtest by 83.7\% and 80.4\%, respectively.

\end{abstract}

\section{Introduction}
\label{sec:intro}

Vision--language models (VLMs) perform well on multimodal question answering and image understanding, but remain prone to hallucination~\citep{liu2024survey, liang2026comprehensive}. This is especially problematic in culturally grounded settings, where models may rely on learned associations instead of visual evidence. For example, a model may associate an Arab country with Islamic art even when the image does not support that interpretation. Detecting this behaviour requires items where the culturally plausible answer and the visually supported answer are separated.

The ImageEval 2026 shared task~\citep{imageeval-2026} focuses on culturally grounded Arabic multimodal understanding. Task~1 includes: 
\begin{itemize}
\item \textit{Subtask 1a} — Spoken VQA: given an image, a spoken question, and three spoken options, predict the correct option.
\item \textit{Subtask 1b} — Hallucination Detection: given an image and three statements, identify the one statement that is visually grounded.
\end{itemize}
We describe team \textit{\textcolor{ForestGreen}{\textit{alkhder}}}'s submission to Subtask~1b in both the English and MSA tracks. Our system uses \texttt{Qwen2.5-VL-7B-Instruct}~\citep{bai2025qwen25vltechnicalreport}, an off-the-shelf 7B vision--language model and relies on zero-shot inference with no training data or fine tuning. We enforce the task constraint that exactly one statement is supported, reducing the problem to a three-way choice. The model selects the statement best supported by the image using next-token log-probabilities. To reduce answer-position bias in multiple-choice prompting~\citep{zheng2024large,pezeshkpour2024large}, we average scores over a Latin-square rotation of the answer slots.

On the English development set, our system obtains a combined accuracy of 0.9500. On the blind test sets, it achieves CI scores of 0.0510 on English and 0.0960 on MSA, ranking 5th and 4th, respectively. These scores improve on the released 3B baselines by $6.1\times$ for English and $5.1\times$ for MSA.
 We analyse the English--MSA performance gap and show that our system has the largest gap among teams appearing in both tracks, which we partly attribute to validating all design choices on English development data only.
The rest of the paper is organised as follows. Section~\ref{sec:background} describes the shared task and related works. Section~\ref{sec:system} presents our method. Section~\ref{sec:setup} details the experimental settings, Section~\ref{sec:results} reports results, and section~\ref{sec:conclusion} concludes.

\section{Background}
\label{sec:background}

\subsection{Task Description and data}

Subtask~1b is a hallucination detection task in which each example contains an image and three statements, only one of which is visually valid. Remaining statements are culturally plausible but unsupported by the image. Systems must label each statement as True if it is supported by the image, and False otherwise.
The dataset includes cases where cultural priors and visual evidence agree, and cases where they conflict. The task is offered both in English and MSA, using the same images and correct answers with translated statements. Although predictions could be copied across tracks, the organisers require each track to be processed independently.
Each item includes an image, three statements, a country label covering one of 18 Arab countries, and a cultural category. Submissions require one CSV row per statement. The data are split into \texttt{train} (3{,}000 items), \texttt{dev} (500), \texttt{devtest} (500), and \texttt{test} (1{,}000). Labels for \texttt{devtest} and \texttt{test} were hidden during evaluation and released after the task closed.
We use no training or fine-tuning. The \texttt{dev} split is used for validation and ablation, and \texttt{test} for the final submission. We do not use \texttt{devtest}, since the development phase had closed when we entered the task. We apply no filtering, augmentation, or resampling, and pass the statements to the model as released.

\subsection{Related Work}

Hallucination in VLMs was first studied in image captioning, where models were shown to describe objects absent from the image \citep{rohrbach2018object}. Later work moved from free-form generation to yes/no probes about image content \citep{li2023evaluating}, a format similar to the official baseline for this subtask. Recent work has also examined whether hallucination evaluations transfer beyond English and Western visual contexts \citep{romero2024cvqa,alwajih2024peacock}. ImageEval 2026 follows this direction, draws items from OASIS \citep{alam2025everydaymmqa}, uses the CFHR metric from the counterfactual hallucination analysis of \citet{mousi-etal-2026-correct}, and includes a spoken track based on \citet{mousi2026said}. We combine likelihood scoring over a fixed answer set with permutation ensembling. Likelihood scoring avoids free-form generation, while permutation ensembling reduces multiple-choice position bias \citep{zheng2024large,pezeshkpour2024large}.

\section{System Overview}
\label{sec:system}

\begin{figure*}[t]
  \centering
  \resizebox{\textwidth}{!}{%
  \begin{tikzpicture}[
      font=\small,
      >={Stealth[length=2.2mm,width=1.7mm]},
      box/.style={draw, rounded corners=2.5pt, align=center, inner sep=5pt,
                  minimum height=13mm, line width=0.6pt},
      io/.style  ={box, fill=maskgrey, draw=cIOd},
      pw/.style  ={box, fill=keepblue, draw=cPWd},
      mc/.style  ={box, fill=cPW, draw=cPWd},
      op/.style  ={box, fill=tem, draw=cOPd},
      flow/.style={->, line width=0.7pt, draw=cLine, rounded corners=2.5pt},
      lbl/.style ={font=\scriptsize, text=cLine, inner sep=1.5pt}
    ]
 
    \node[io] (input) {image $I$\\[1pt] $t_0,\,t_1,\,t_2$};
 
    \node[pw, right=10mm of input, yshift=10mm] (pw)
      {\textbf{pointwise view}\\[1pt] each $t_i$ scored alone\\[1pt]
       {\scriptsize 3 fwd.\ passes}};
    \node[mc, right=10mm of input, yshift=-10mm] (mc)
      {\textbf{multiple-choice view}\\[1pt] $\times 3$ rotated orders\\[1pt]
       {\scriptsize 3 fwd.\ passes}};
 
    \coordinate (mid) at ($(pw.east)!0.5!(mc.east)$);
    \node[op, right=10mm of mid, anchor=west] (pool)
      {\textsc{LogSumExp} pooling\\[1pt]
       $\rightarrow s^{\mathrm{PW}}_i,\ s^{\mathrm{MC}}_i$};
    \node[op, right=8mm of pool] (fuse) {log-softmax\\[1pt] fuse $1{:}1$};
    \node[io, right=8mm of fuse]  (out)
      {$\hat{y}=\arg\max_i s_i$\\[1pt] {\scriptsize one-true label}};
 
    \coordinate (j) at ([xshift=5.5mm]pw.east |- pool);
    \draw[flow]   (input.east) -- ++(5mm,0)   |- (pw.west);
    \draw[flow]   (input.east) -- ++(5mm,0)   |- (mc.west);
    \draw[flow,-] (pw.east)    -- ++(5.5mm,0) |- (j);
    \draw[flow,-] (mc.east)    -- ++(5.5mm,0) |- (j);
    \draw[flow]   (j)    -- (pool.west);
    \draw[flow]   (pool) -- (fuse);
    \draw[flow]   (fuse) -- (out);
 
    \begin{scope}[on background layer]
      \node[draw=cIOd!60, dashed, rounded corners=3pt, fill=cBG,
            inner sep=3.5mm, fit=(pw)(mc)] (group) {};
    \end{scope}
    \node[lbl, below=0.8mm of group] {6 forward passes per item};
 
  \end{tikzpicture}}
  \caption{System overview.}
  \label{fig:pipeline}
\end{figure*}
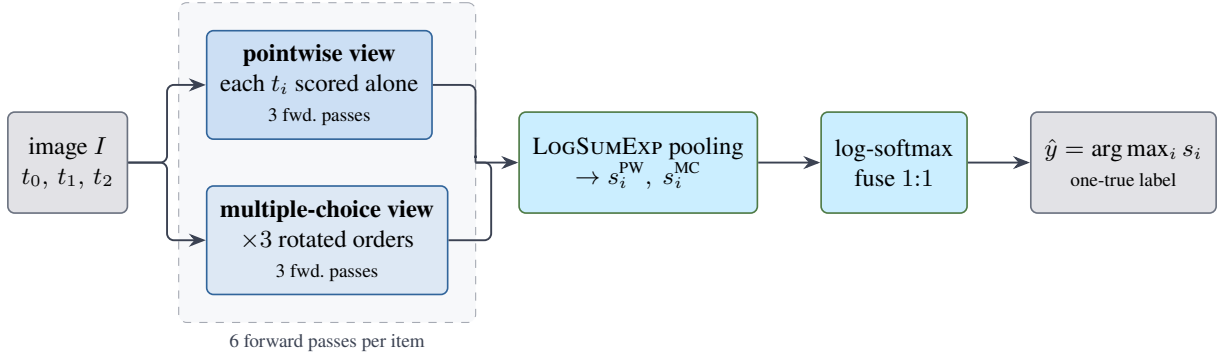
 
Let an item be an image $I$ with statements $t_0, t_1, t_2$. The system (Figure~\ref{fig:pipeline}) produces a single winner index $\hat{y}$ and emits \texttt{true} for $\hat{y}$, \texttt{false} otherwise.

\subsection{Logit-level answer reading}
\label{sec:logit}

 Instead of using \texttt{generate()}, we run a forward pass and read the answer from the next-token probability distribution. To reduce tokenisation effects, each label is represented by a group of surface forms. For example, $\mathcal{T} = \{$\texttt{True}, \texttt{true}, \texttt{\_True}, \texttt{\_true}, \texttt{TRUE}$\}$ and analogously $\mathcal{F}$ for False, where \texttt{\_} denotes a leading space. We pool each group using \texttt{logsumexp}:
\begin{equation}
\ell(\mathcal{S} \mid x) =
\log \sum_{s \in \mathcal{S}} P(s \mid x).
\end{equation}

This avoids parse failures and the system does not have to interpret generated text or fall back to a default label. It also preserves confidence, since a weak or strong preference for \texttt{True} remain distinguishable. In practice, we observe that this is faster than generating and parsing free-form text, allowing us to use a larger model.

\subsection{Scoring}

We score each item in two ways.

\paragraph{Pointwise.} The model sees $I$ and one statement $t_i$, and is instructed to judge only what the image shows. The score is a log-odds margin:
\begin{equation}
s^{\text{PW}}_i = \ell(\mathcal{T} \mid I, t_i) - \ell(\mathcal{F} \mid I, t_i).
\end{equation}

\paragraph{Multiple-choice.}
The model sees $I$ and all three statements together with an explicit statement of the constraint (``exactly ONE statement is supported''), and must answer with the corresponding option number. The score is the pooled log-probability of the corresponding option token allowing the model compare candidates simultaneously instead of three independent yes/no decisions.

\subsection{Rotation averaging}
\label{sec:rotation}

To reduce answer-position bias, we score the multiple-choice prompt under three cyclic permutations,
$\Pi = \{(0,1,2),(1,2,0),(2,0,1)\}$, so that each statement appears once in each slot. For each original statement, we average the score of the slot it occupies:
\begin{equation}
s^{\text{MC}}_i =
\frac{1}{|\Pi|}
\sum_{\pi \in \Pi}
\ell\big(\mathcal{O}_{\mathrm{pos}_\pi(i)} \mid I, \pi(t_{0:2})\big),
\end{equation}
where $\mathcal{O}_k$ is the token group for option $k$. Since every statement appears once in every slot, any fixed preference for a particular answer position contributes equally to all statements.

\subsection{Fusion and decoding}
 
Each score is normalised into a distribution over the three statements and combined:
\begin{align}
\small
z_i &= W_{\text{PW}}\,\text{log\_softmax}(s^{\text{PW}})_i \nonumber
    + W_{\text{MC}}\,\text{log\_softmax}(s^{\text{MC}})_i, \\
\hat{y} &= \arg\max_i z_i .
\end{align}
For the submission, we set $W_{\text{PW}} = W_{\text{MC}} = 1.0$. Total cost amounts to six forward passes per item (three pointwise, three rotations). Decoding is deterministic with no sampling and no temperature.
\section{Experimental Setup}
\label{sec:setup}
 
\subsection{Model and implementation} We use \texttt{Qwen/Qwen2.5-VL-7B-Instruct}~\citep{bai2025qwen25vltechnicalreport} via HuggingFace Transformers. The model is used without fine-tuning in \texttt{bf16} without quantisation. The vision encoder is capped at \texttt{MAX\_PIXELS} $=1024\times28\times28$, on an NVIDIA L4 (24\,GB) on Google Colab Pro. Precision and pixel cap were selected automatically from the detected GPU (Appendix~\ref{app:hparams}). Throughput was 4.17\,s/item (EN) and 4.18 (MSA) giving 69.5 minutes per 1{,}000-item split. Predictions were checkpointed per item to JSONL, making runs resumable. Images were downloaded with four parallel
workers. 

\subsection{Weight selection} 
We tuned the relative weight of the two scores.  Sweeping $W_{\text{PW}}$ on English \texttt{dev} with $W_{\text{MC}}$ fixed at $1.0$ placed the optimum at $W_{\text{PW}} = 0.25$, giving a combined accuracy of $0.9520$ against $0.9500$ at the untuned $1.0$ setting for one item in 500. We judged it indistinguishable
from noise on a split of this size and submitted the 1:1 configuration. The 0.25 optimum corresponds to a single item, the 95\% binomial interval on $n{=}500$ spans $\pm 1.9$ points, and fitting $W_{\text{PW}}$ to that margin would tune split-specific noise. 

\subsection{Split usage} We used \texttt{dev} (500 items, labelled) for
validation, ablation, and weight sweep, and \texttt{test} (1{,}000 items, unlabelled at submission time) for the official runs.  The \texttt{train}
split was not used, since the system involves no training, and \texttt{devtest} was not used because the development phase had closed when we entered. Each track was run end to end on its own inputs, producing two submissions: Codabench 883496 for English and 17050 for MSA.

\subsection{Evaluation Metrics} Let $Q^+$ denote the grounded statement and $Q^-$ the hallucinated statements. The official ranking metric is Contrastive Instability, which measures, among items with at least one correct label, the share of items that are not fully correct. Secondary metrics are combined accuracy (all three labels correct), CFHR (among items whose $Q^+$ was found, the share that still affirmed a hallucination), and $Q^+$/$Q^-$ accuracy.

 \section{Results}
\label{sec:results}

\subsection{System Performance}

Table~\ref{tab:official} reports our official test-set
figures against the released baselines and the top-ranked
system on each track. Both submissions are single runs of the
configuration described so far.  

On the English track we finished fifth with a CI of $0.0510$, and on the MSA track fourth with a CI of $0.0960$. In absolute terms, our system labelled all three statements correctly on 949 of 1{,}000 English items and on 904 MSA items. Against the released 3B baselines our system reduces CI by a factor of 6.1 on English and 5.1 on MSA, although the baselines are reported on \texttt{devtest} rather than \texttt{test} and the comparison is mainly indicative.

CFHR is $0.0000$ for our system and all entries on both leaderboards.  This is attributed to the one-true decoding constraint, under which several official metrics become deterministic functions of three-way selection accuracy: $\text{CI}=1-\text{acc}$, $\text{CFHR}=0$, and $Q^-\text{acc}=(1+\text{acc})/2$; see Appendix~\ref{app:derivation}.

\begin{table}[t]
\centering
\footnotesize
\setlength{\tabcolsep}{4pt}
\begin{tabular}{@{}lrrrr@{}}
\toprule
System & CI\,$\downarrow$ & Comb.\,$\uparrow$ & $Q^{+}\uparrow$ & $Q^{-}\uparrow$ \\
\midrule
\multicolumn{5}{@{}l}{\textit{English track} (\texttt{test}, $n{=}1000$)} \\
Baseline 3B$^{\dagger}$     & 0.3133 & 0.6840 & 0.9400 & 0.8390 \\
Rank 1 (\texttt{miles\_ma}) & 0.0290 & 0.9710 & 0.9710 & 0.9855 \\
Ours (rank 5)               & \textbf{0.0510} & 0.9490 & 0.9490 & 0.9745 \\
\addlinespace
\multicolumn{5}{@{}l}{\textit{MSA track} (\texttt{test}, $n{=}1000$)} \\
Baseline 3B$^{\dagger}$       & 0.4900 & 0.5080 & 0.9400 & 0.8390 \\
Rank 1 (\texttt{ahmedbakr})   & 0.0360 & 0.9640 & 0.9640 & 0.9820 \\
Ours (rank 4)                 & \textbf{0.0960} & 0.9040 & 0.9040 & 0.9520 \\
\bottomrule
\end{tabular}
\caption{Official Subtask~1b results. $^{\dagger}$Baseline figures are
reported on \texttt{devtest}.}
\label{tab:official}
\end{table}

\subsection{Ablation Study}

Table~\ref{tab:ablation} reports pre-submission ablations on English \texttt{dev}. Differences between settings are small, corresponding to one to three items.

\begin{table}[h]
\centering
\footnotesize
\begin{tabular}{lrr}
\toprule
Configuration & Comb.\ acc & CI\,$\downarrow$ \\
\midrule
Pointwise only & 0.9440 & 0.0560 \\
Multiple-choice only & 0.9500 & 0.0500 \\
Both, 1:1 (\textbf{submitted}) & 0.9500 & 0.0500 \\
Both, swept $W_{\text{PW}}{=}0.25$ & 0.9520 & 0.0480 \\
\bottomrule
\end{tabular}
\caption{View ablation on English \texttt{dev} ($n{=}500$).}
\label{tab:ablation}
\end{table}

Pointwise scoring reaches 0.9440 while multiple-choice scoring reaches 0.9500, three items higher on the 500-item split. The two are therefore nearly interchangeable, with a small advantage for multiple-choice scoring.

Equal-weight fusion also reaches 0.9500, matching but not improving on the multiple-choice view. Sweeping $W_{\text{PW}}$ raises accuracy to 0.9520 at $W_{\text{PW}}=0.25$, a gain of one item. Since we did not run the pipeline on the 3B baseline model, we cannot separate the effects of the one-true constraint, logit-level scoring, and model scale.

Although equal-weight fusion matches multiple-choice scoring on dev, we
opt for the pointwise as none of the fusion settings we tested performed worse than using the multiple-choice alone. Moreover, the pointwise judges each statement independently and does not necessarily degrade when the hallucinated statements are semantically close to the image.

\subsection{English--MSA Gap}
\label{sec:tracks}

Our English and MSA submissions differ by 4.5 points of combined accuracy. Table~\ref{tab:gap} compares this with the other submissions that entered both tracks. Our gap is the largest in the field, while NYUAD's 0.6-point gap suggests that the backbone alone does not explain a deficit of this size.

\begin{table}[h]
\centering
\small
\begin{tabular}{lrrr}
\toprule
Team & EN & MSA & $\Delta$ \\
\midrule
NYUAD & 0.9590 & 0.9530 & 0.6 \\
myamin66 & 0.9670 & 0.9440 & 2.3 \\
keslerady & 0.8870 & 0.8580 & 2.9 \\
md\_faisal & 0.9330 & 0.8990 & 3.4 \\
\textbf{\textcolor{ForestGreen}{\textit{alkhder}} (ours)} & 0.9490 & 0.9040 & \textbf{4.5} \\
\bottomrule
\end{tabular}
\caption{Combined accuracy per track and the gap in points, for teams appearing on both EN and MSA tracks.}
\label{tab:gap}
\end{table}

We attribute the gap mainly to procedure. Prompt wording, answer token groups, and score weights were all validated on English \texttt{dev} and then transferred to MSA without Arabic-side tuning. Closing this gap would improve the MSA result.

The prediction distributions indicate that the two tracks were run independently. English produced $\{0{:}324, 1{:}352, 2{:}324\}$, while MSA produced $\{0{:}324, 1{:}341, 2{:}335\}$. Both distributions are close to uniform, suggesting little residual slot preference after rotation.

\subsection{Dev-to-Test Transfer}
\label{sec:transfer}

English \texttt{dev} accuracy was 0.9500, and the blind test result was 0.9490, one item per thousand apart. This close match is consistent with the system having no fitted parameters beyond choices made on \texttt{dev}. Leaderboard ranks should be interpreted with some caution. On English, a binomial 95\% interval around our score is approximately $\pm 1.4$ points, so nearby ranks are not clearly separated by aggregate scores alone. The gap to the top English system is larger, while the 60-item gap to the top MSA system is unambiguous.

\subsection{Error Analysis}
\label{sec:error-analysis}
The following is post-submission analysis, using test labels released after the evaluation period closed. No configuration was changed and no inference was re-run using these labels. 
\begin{table}[htbp!]
\centering
\small
\begin{tabular}{lrrr}
\toprule
Category & $n$ & EN acc. & MSA acc. \\
\midrule
Objects \& Clothing & 94 & 0.9681 & 0.9043 \\
Vehicles \& Transport & 148 & 0.9662 & 0.9189 \\
People \& Society & 185 & 0.9568 & 0.9351 \\
Culture \& Arts & 175 & 0.9543 & 0.9029 \\
Buildings \& Landmarks & 53 & 0.9434 & 0.9434 \\
History \& Identity & 64 & 0.9375 & 0.9219 \\
Food \& Cooking & 93 & 0.9355 & 0.8710 \\
Religion & 87 & 0.9310 & 0.8851 \\
Sports \& Rec. & 101 & 0.9208 & 0.8416 \\
\bottomrule
\end{tabular}
\caption{Combined accuracy by cultural category.}
\label{tab:category}
\end{table}

\begin{table*}[t!]
\centering
\footnotesize
\begin{tabular}{lrll}
\toprule
Country & $n$ & EN acc. [95\% CI] & MSA acc. [95\% CI] \\
\midrule
Morocco & 85 & 0.9882 [0.9363, 0.9979] & 0.9412 [0.8696, 0.9746] \\
UAE & 81 & 0.9877 [0.9333, 0.9978] & 0.9136 [0.8322, 0.9575] \\
Sudan & 80 & 0.9875 [0.9325, 0.9978] & 0.9125 [0.8302, 0.9570] \\
Palestine & 84 & 0.9643 [0.9002, 0.9878] & 0.8929 [0.8088, 0.9426] \\
Kuwait & 91 & 0.9560 [0.8924, 0.9828] & 0.9231 [0.8496, 0.9622] \\
Syria & 83 & 0.9518 [0.8825, 0.9811] & 0.8795 [0.7922, 0.9332] \\
Algeria & 87 & 0.9425 [0.8724, 0.9752] & 0.9310 [0.8576, 0.9680] \\
Bahrain & 98 & 0.9388 [0.8728, 0.9716] & 0.8878 [0.8101, 0.9362] \\
Yemen & 78 & 0.9359 [0.8586, 0.9723] & 0.8974 [0.8105, 0.9471] \\
Lebanon & 76 & 0.9342 [0.8551, 0.9716] & 0.9211 [0.8383, 0.9633] \\
Tunisia & 44 & 0.9318 [0.8177, 0.9765] & 0.8409 [0.7063, 0.9207] \\
Iraq & 85 & 0.9176 [0.8396, 0.9595] & 0.8941 [0.8109, 0.9433] \\
Saudi Arabia & 28 & 0.7857 [0.6046, 0.8979] & 0.8571 [0.6851, 0.9430] \\
\bottomrule
\end{tabular}
\caption{Combined accuracy by country of origin with  95\% confidence intervals.}
\label{tab:country}
\end{table*}

Accuracy varies from  0.84 to 0.97 across the nine cultural categories for both English and MSA (Table~\ref{tab:category}), though a $\chi^2$ test of category against correctness is not significant at $p=0.784$ (EN) / $p=0.287$ (MSA), consistent with error counts too small per cell to establish category-level effects at $n=1000$.

Table~\ref{tab:country} reports country breakdown. English and MSA predictions are paired per item, giving both-correct/EN-only/MSA-only/both-wrong counts of 897/52/7/44. McNemar's exact test on discordant cells rejects symmetry ($p < 0.0001$), and a paired bootstrap (10{,}000 resamples) puts the EN$-$MSA accuracy gap at 0.0450 (95\% CI [0.0310, 0.0600]). Thus, the EN advantage is unlikely to be noise. Of the 44 items missed by both tracks, 34 selected the same wrong statement and 10 selected different ones, suggesting both shared distractor (false) effects and track-specific errors. 

\subsection{Qualitative Examples}
Appendix~\ref{sec:examples},  Table~\ref{tab:examples} shows one success and two failures, manually checked against released labels and images. For a rice-and-meat dish, the system rates almost as compatible with ``Indian cuisine'' (4.62) as with the grounded `Middle Eastern cuisine'' (5.12) but the multiple-choice score separates them.
The success case shows the same effect culturally, patterned rugs at an Arab gathering are expected to serve as guest seating, and the model rates that statement as plausible, yet the multiple-choice selects the grounded ``decorative'' reading. The shared failure is a scene-stereotype error in which both tracks read soldiers around a wounded person as engaging in combat''. The MSA-only failure reflects translation noise as the model chooses baptism'' over the grounded foot-washing ritual, whose MSA gold statement was mistranslated as strange religious rituals''. 

 \section{Conclusion}
\label{sec:conclusion}

We described team alkhder's submission to Subtask~1b of ImageEval 2026 in the English and MSA tracks. Our system uses a pre-trained 7B vision--language model in a zero-shot setting, with no task-specific training or fine-tuning. It exploits the task constraint that exactly one statement is grounded, reads answers from next-token log-probabilities, and reduces position bias through rotation averaging. Our main contribution is an inference and prompting strategy. The system achieved CI scores of 0.0510 on English and 0.0960 on MSA, ranking 5th and 4th, respectively.  

Future work should validate prompts and answer-token groups directly on Arabic, and compare baseline sizes to separate the effects of model scale from the constrained formulation.

\section*{Limitations}

Results are based on one underlying model, prompt family, and run per track. Decoding is deterministic, but we did not measure prompt-paraphrase variance. All validation was conducted on 500 English \texttt{dev} items, with no Arabic-side validation, which likely contributed to the English--MSA gap.

The comparison with the official baselines is indicative since the baselines are reported on \texttt{devtest} and our results on \texttt{test}. Cross-team comparisons use published aggregate scores only, without per-item predictions. Finally, the constrained formulation relies on the dataset guarantee that exactly one statement is grounded, and does not directly apply to open-ended hallucination detection.
 
\bibliography{custom}
\appendix
 
\section{Hyperparameters and Configuration}
\label{app:hparams}
 
\begin{table}[h]
\centering
\small
\begin{tabular}{ll}
\toprule
Setting & Value \\
\midrule
Model & \texttt{Qwen2.5-VL-7B-Instruct} \\
Precision & \texttt{bf16} (not quantised) \\
\texttt{MAX\_PIXELS} & $1024\times28\times28$ \\
Training & none (zero-shot) \\
Views & pointwise $+$ multiple-choice \\
Rotations & $(0,1,2),(1,2,0),(2,0,1)$ \\
Fwd.\ passes / item & 6 \\
Sampling & none (deterministic) \\
$W_{\text{PW}}, W_{\text{MC}}$ & 1.0, 1.0 \\
True tokens & \texttt{True/true/\_True/\_true/TRUE} \\
False tokens & \texttt{False/false/\_False/...} \\
Option tokens & \texttt{0,1,2} ($\pm$ leading space) \\
Hardware & NVIDIA L4, 24\,GB \\
Throughput & 4.17\,s/item (EN), 4.18 (MSA) \\
\bottomrule
\end{tabular}
\caption{Submitted system configuration.}
\end{table}
 
Precision and resolution were selected automatically from the detected GPU: A100 (40\,GB) $\rightarrow$ bf16, $1280\times28\times28$; L4 (24\,GB) $\rightarrow$ bf16, $1024\times28\times28$ (used); T4 (16\,GB) $\rightarrow$ 4-bit NF4, $512\times28\times28$.
\section{Qualitative Examples}\label{sec:examples}
\begin{table*}[t]
\centering
\footnotesize
\setlength{\tabcolsep}{4pt}
\begin{tabular}{@{}p{0.13\linewidth}p{0.33\linewidth}p{0.33\linewidth}c@{}}
\toprule
\textbf{Item} & \textbf{Grounded statement (gold)} & \textbf{Model selection} & \textbf{Result} \\
\midrule
UAE \newline \texttt{c8cfb559} \newline \textit{Traditional Cuisines} &
``The dish shown in the image is likely Middle Eastern cuisine.''
(distractor: Italian, Indian) &
same as gold statement & \checkmark\ (EN \& MSA) \\
\addlinespace
Lebanon \newline \texttt{4ff7e8f3} \newline \textit{Rituals \& Ceremonies} &
``The patterned rugs are likely used to provide a decorative cultural
element for the gathering.''
(culturally expected distractor: ``seating area for guests'') &
same as gold statement & \checkmark\ (EN \& MSA) \\
\addlinespace
Iraq \newline \texttt{c409e355} \newline \textit{Healthcare \& Well-being} &
``The primary focus of the soldiers in the image is providing medical
assistance.'' &
``The primary focus of the soldiers in the image is engaging in
combat.'' & \texttimes\ (EN \& MSA) \\
\addlinespace
Syria \newline \texttt{7395c814} \newline \textit{Heritage \& History} &
``The occasion depicted in the image is a religious foot-washing
ritual.'' \emph{(MSA gold mistranslated as ``strange religious
rituals'')} &
``The occasion depicted in the image is a baptism event.'' &
\checkmark\ EN, \texttimes\ MSA \\
\bottomrule
\end{tabular}
\caption{Qualitative examples from the released \texttt{test} labels,
verified against the images. Statements are quoted from the released
English data; the MSA track used the released Arabic translations.}
\label{tab:examples}
\end{table*}
\newpage 
\section{Prompts}
\label{app:prompts}

\paragraph{Pointwise.}
\begin{quote}\small\ttfamily
You are checking a statement against an image for visual hallucination. Judge only from what the image actually shows, not from what sounds plausible.\\[0.3em]
Statement: "\{statement\}"\\[0.3em]
If the image clearly supports the statement, answer True. If it is not shown in the image or is contradicted by it, answer False. Answer with exactly one word: True or False.
\end{quote}
 
\paragraph{Multiple-choice.}
\begin{quote}\small\ttfamily
Look carefully at the image. Below are three statements about it. Exactly ONE of them is actually supported by what the image shows; the other two are plausible-sounding but are NOT supported by the image.\\[0.3em]
0. \{s0\}\quad 1. \{s1\}\quad 2. \{s2\}\\[0.3em]
Which statement is the one actually supported by the image? Answer with a single digit: 0, 1, or 2.
\end{quote}
\paragraph{MSA track prompts.} The MSA track used the same two templates verbatim. Only the statement slots differ: $\{statement\}$ and $\{s_0\}$, $\{s_1\}$, $\{s_2\}$ are filled with the track's Arabic statements, passed to the model exactly as released. The English instruction wrapper, the answer token groups, and the rotation scheme were identical across tracks. No Arabic-language instruction wording, answer tokens (e.g. \foreignlanguage{arabic}{خطأ}/\foreignlanguage{arabic}{صح}), and no translation step were used.
 

\section{Metric Derivation}
\label{app:derivation}

Let $n$ be the number of items, and let $c$ be the number for which the grounded statement is selected. Under one-true decoding, each item receives exactly one \texttt{true} label and two \texttt{false} labels.

If the grounded statement is selected, all three labels are correct. If it is not selected, exactly one label is still correct and one of the two hallucinated statements remains labelled \texttt{false}. Therefore every item has at least one correct label, so the CI denominator is $n$, and
\begin{equation}
\text{CI} = \frac{n-c}{n} = 1 - \text{acc}.
\end{equation}

For items where $Q^+$ is correctly selected, both hallucinated statements are labelled \texttt{false}. The system therefore never affirms a hallucination after selecting the grounded statement, giving
\begin{equation}
\text{CFHR} = 0.
\end{equation}

Finally, there are $2n$ hallucinated statements in total. Correctly solved items contribute two correct $Q^-$ labels, while incorrectly solved items contribute one. Thus
\begin{equation}
Q^-\text{acc}
=
\frac{2c + (n-c)}{2n}
=
\frac{1+\text{acc}}{2}.
\end{equation}

For example, with $c=949$ and $n=1000$, this gives $Q^-\text{acc}=0.9745$, matching the official scorer.




\end{document}